\newif\ifarxiv
\newif\ifralfinal

\ralfinaltrue \arxivtrue
\documentclass[letterpaper, 10 pt, journal, twoside]{IEEEtran}  %

\IEEEoverridecommandlockouts                              %

\usepackage{graphicx}
\usepackage[table,xcdraw]{xcolor}
\usepackage{amsmath}
\usepackage{mathtools}
\usepackage{booktabs}
\usepackage{capt-of}
\usepackage{eso-pic}
\usepackage{hyperref}

\usepackage{makecell}
\usepackage{booktabs}
\usepackage{comment}
\usepackage{url} 
\usepackage{amssymb}
\usepackage{bm}
\usepackage{xspace}

\usepackage{tabularx}
\usepackage{multirow}
\usepackage{microtype}

\ifarxiv
\AddToShipoutPicture*{%
     \AtTextUpperLeft{%
         \put(-3.5,10){
           \begin{minipage}{\textwidth}
              \scriptsize
              Preprint version; final version available at \url{https://doi.org/10.1109/LRA.2022.3187836}
           \end{minipage}}%
     }%
}

\fi

\ifralfinal
\fi

\makeatletter
\DeclareRobustCommand\onedot{\futurelet\@let@token\@onedot}
\def\@onedot{\ifx\@let@token.\else.\null\fi\xspace}

\def\eg{\emph{e.g}\onedot} 
\def\ie{\emph{i.e}\onedot}

\def\etal{\emph{et al}\onedot}
\makeatother

\begin{document}

\title{Point Label Aware Superpixels for Multi-species Segmentation of Underwater Imagery}
\bstctlcite{IEEEexample:BSTcontrol}

\author{Scarlett Raine$^{1,2}$, Ross Marchant$^{1,2}$, Brano Kusy$^{2}$, Frederic Maire$^{1}$ and Tobias Fischer$^{1}$%
\ifralfinal
\thanks{Manuscript received: February 24, 2022; Revised May 24, 2022; Accepted June 16, 2022.}%
\thanks{This letter was recommended for publication by Editor Youngjin Choi upon evaluation of the reviewers' comments. This research was partially supported by the QUT Centre for Robotics. TF was supported by the Australian Government, via grant AUSMURIB000001 associated with ONR MURI grant N00014-19-1-2571, funding from ARC Laureate Fellowship FL210100156, and Intel's Neuromorphic Computing Lab (\textit{Corresponding author: Scarlett Raine.} {\tt\footnotesize scarlett.raine@hdr.qut.edu.au})} %
\fi
\thanks{$^{1}$Scarlett Raine, Ross Marchant, Frederic Maire and Tobias Fischer are with the QUT Centre for Robotics and School of Electrical Engineering and Robotics, Queensland University of Technology, Brisbane, Australia.}%
\thanks{$^{2}$Scarlett Raine, Ross Marchant and Brano Kusy are with Data61, CSIRO, Brisbane, Australia.}%
\ifralfinal
\thanks{Digital Object Identifier (DOI): \url{https://doi.org/10.1109/LRA.2022.3187836}.}
\fi
}
\pagenumbering{gobble}
\maketitle
\begin{abstract}
    Monitoring coral reefs using underwater vehicles increases the range of marine surveys and availability of historical ecological data by collecting significant quantities of images. Analysis of this imagery can be automated using a model trained to perform semantic segmentation, however it is too costly and time-consuming to densely label images for training supervised models.  In this letter, we leverage photo-quadrat imagery labeled by ecologists with sparse point labels.  We propose a point label aware method for propagating labels within superpixel regions to obtain augmented ground truth for training a semantic segmentation model.  Our point label aware superpixel method utilizes the sparse point labels, and clusters pixels using learned features to accurately generate single-species segments in cluttered, complex coral images.  Our method outperforms prior methods on the UCSD Mosaics dataset by 3.62\% for pixel accuracy and 8.35\% for mean IoU for the label propagation task, while reducing computation time reported by previous approaches by 76\%.  We train a DeepLabv3+ architecture and outperform state-of-the-art for semantic segmentation by 2.91\% for pixel accuracy and 9.65\% for mean IoU on the UCSD Mosaics dataset and by 4.19\% for pixel accuracy and 14.32\% mean IoU for the Eilat dataset. 
\end{abstract}

\ifralfinal
\fi

\section{Introduction}
\label{sec:intro}  

\ifralfinal
\IEEEPARstart{M}{arine}
\else
Marine
\fi surveys aim to identify and monitor reef health changes, and are traditionally completed manually by ecologists that collect photographs along transects and subdivide or crop the images to 1m by 1m quadrats for further analysis \cite{obura2019coral}. Domain experts then label randomly distributed pixels in the photo-quadrats into taxonomic or morphological classes \cite{murphy2010observational}. These manually labeled pixels, sparsely distributed in the image, are called \textit{point labels}, and offer a middle ground between the weak training signal of image-level labels and the intensive, time-consuming, and costly method of densely labeling every pixel in each image (\textit{dense segmentation}) \cite{bearman2016s}. Such efficient labeling is critical given the recent advances in broad-scale marine survey methods using autonomous underwater vehicles that have led to an increasing range and accuracy of survey data \cite{sward2019systematic, xu2019deep, monk2018marine}.      

However, point-based sparse annotations cannot indicate growth or shrinkage of coral communities, and cannot be used to analyze spatial distributions at a fine scale \cite{pavoni2021challenges}. Also, other reef health indicators such as composition of coral species, biological diversity and potential for recovery \cite{obura2019coral} can be best obtained from dense segmentations. Therefore, automatically obtaining dense segmentations from these point labels is desirable.

\begin{figure}[t]
\includegraphics[width=\columnwidth, clip, trim=3.5cm 2cm 3.5cm 1cm]{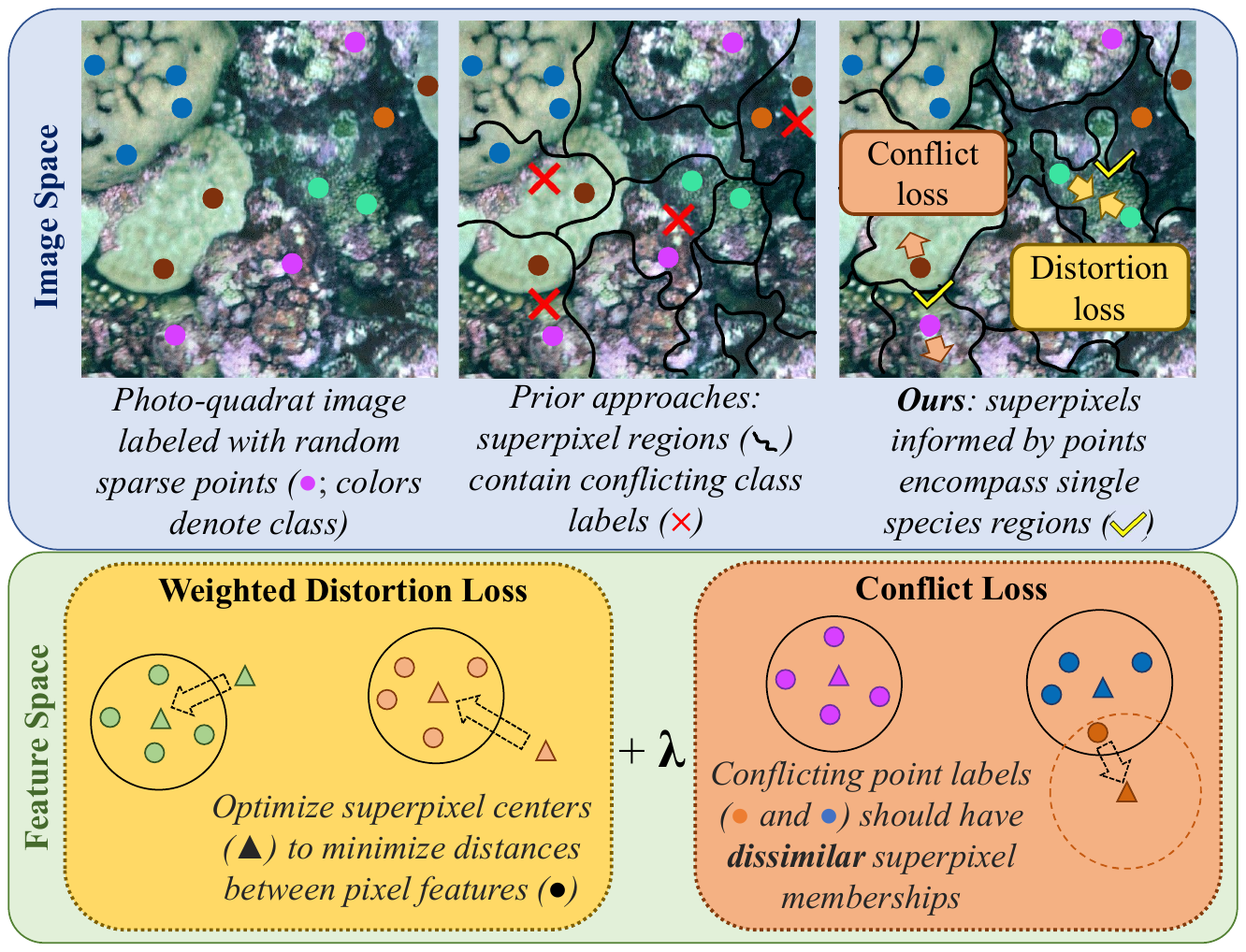}
\caption{Our point label aware approach to superpixels creates segments in coral images which closely conform to complex boundaries and leverage sparse point labels provided by domain experts.  \textbf{Top: }Prior approaches use color and spatial information to create superpixels, whereas our method leverages the locations and classes of point labels to generate single species regions. \textbf{Bottom:} Our novel loss function is comprised of two terms -- the distortion loss groups pixels with similar features while the conflict loss ensures conflicting class labels have dissimilar superpixel memberships.}
\label{fig:frontpage}
\end{figure}

More specifically, the aim of the semantic segmentation task is to classify every pixel in a query image into one of a number of predefined classes \cite{long2015fully}.  When a neural network performs semantic segmentation, the network is typically trained on densely labeled data in the form of image-mask pairs, where every pixel in the training image is assigned a class label.  
This letter introduces a novel method for propagating the random point labels that domain experts provide to obtain an augmented ground truth for training neural networks to perform segmentation of underwater imagery (Fig.~\ref{fig:frontpage}).

Previous methods have propagated point labels using superpixel algorithms, which cluster pixels into segmented regions called superpixels. A superpixel groups perceptually similar and spatially connected pixels, and should ideally adhere closely to object boundaries \cite{achanta2012slic}. Prior approaches use the SLIC \cite{achanta2012slic} or SEEDs \cite{van2012seeds} algorithms, which cluster pixels based on color information \cite{alonso2018semantic, alonso2019coralseg, yu2019fast}.  The superpixels are then used to propagate the point labels to obtain an augmented ground truth; either by layering multiple superpixel decompositions \cite{alonso2018semantic}  or by selecting additional pixels from each segment to add to the ground truth \cite{yu2019fast}.

However, we argue that using color information for coral image segmentation is suboptimal, as coral images are subject to changes in lighting and coloration due to the water column, and coral instances are often poorly defined due to the fractal-like nature of their boundaries, intricacy of growth, and overlapping specimens.  We instead propose using deep features from a neural network to improve superpixel generation in the coral context (Fig.~\ref{fig:frontpage}). 

In this letter, we present the following contributions to the coral segmentation problem:
\begin{enumerate}
    \item We propose a novel method of propagating sparse point labels to obtain augmented ground truth masks for coral images. The underlying feature extractor does not need to be re-trained for new species, and thus we can generate ground truth masks for previously unseen coral species. As our approach does not rely on multi-level operations, it is significantly faster than prior methods.
    \item We introduce a point label aware loss function that generates superpixels that encompass single-species regions (using a conflict loss term) and conform to complex coral boundaries (using a distortion loss term). 
    \item We design an ensemble method that combines superpixel segments from multiple classifiers, thus reducing unlabeled regions in the augmented ground truth and improving the robustness of superpixels. 
    \item We demonstrate significant improvements in label propagation on the UCSD Mosaics dataset (3.6\% in pixel accuracy and 8.4\% in mIoU). We further show that DeepLabv3+ models trained on our augmented ground truth masks outperform prior approaches on the UCSD and Eilat datasets by 9.7\% and 14.3\% mIoU respectively.
\end{enumerate}

\noindent We have made our code available to foster future research in this area: \url{https://github.com/sgraine/point-label-aware-superpixels}.
\section{Related Work}
\label{sec:related}
Automating the analysis of underwater imagery is an active field of research at the intersection of computer vision, marine biology, and robotics \cite{alonso2019coralseg, pavoni2021taglab, song2021development}.  Recently, there have been advances in the use of remotely operated vehicles \cite{sward2019systematic} and autonomous underwater vehicles \cite{dayoub2015robotic, bonin-font, shields2020towards} for performing marine surveys, mapping species of interest, and managing predators such as the Crown of Thorns starfish.  Some robotic platforms have used deep learning for habitat or image classification on-board \cite{shields2020towards, gonzalez2020monitoring}.  While Arain \etal used semantic segmentation for obstacle detection \cite{arain2019improving}, segmentation has not yet been widely used for on-board multi-species analysis of underwater imagery.  This section reviews advances in the semantic segmentation of underwater imagery, weakly supervised approaches for point labels, and methods for augmenting point labels using superpixel algorithms.

\vspace*{-0.1cm}
\subsection{Segmentation of Underwater Imagery}
\label{subsec:underwater_seg}
There are many approaches for \emph{classification} of underwater images and patches~\cite{gonzalez2020monitoring, gomez2019towards, raine2020multi, mahmood2018deep, modasshir2018mdnet, gomez2019coral, chen2021new}.  Modasshir \etal~\cite{modasshir2018coral} describe their approach for detection and counting of corals using an AUV. \cite{modasshir2020enhancing} builds atop of ~\cite{modasshir2018coral} by employing a semi-supervised approach to bounding box detection of targets that leverages temporal information to extract additional examples for re-training their detector.  

There are comparatively fewer approaches for semantic \emph{segmentation} of underwater imagery \cite{pavoni2021challenges, song2021development, liu2020semantic}. These semantic approaches typically use DeepLab \cite{chen2018encoder}, a popular segmentation network, which has outperformed other architectures (often with some modifications and extensions). DeepLab and other segmentation approaches require full supervision during training, necessitating the curation and labeling of custom, densely labeled datasets. Towards this goal, TagLab speeds up \emph{dense} annotation of large orthoimages through automated tools \cite{pavoni2021taglab}. However, a fully automated method that leverages the existing, large quantities of \emph{sparse} point-labeled imagery without additional human labor is desirable for large-scale reef monitoring.      
\vspace*{-0.1cm}
\subsection{Weakly Supervised Approaches for Point Labels}
\label{subsec:superpixels}

Training models on sparse or weakly labeled data to perform dense semantic segmentation is an active area of research.  Many approaches focus on scribble annotations \cite{vernaza2017learning, hua2021semantic}, where images are labeled by drawing paths over regions in the frame.  Training a model from sparse points takes this problem a step further by only requiring sparse point labels in training images as opposed to scribble annotations.  For semantic segmentation of remote sensing images \cite{hua2021semantic} or objects in scenes \cite{bearman2016s}, the point labels are assumed to be placed centrally within the objects of interest.  In ecology, the sparse point labels are randomly distributed in an image, meaning that methods designed to exploit ``objectness" \cite{bearman2016s} cannot be applied. Instead, label propagation approaches have been favored in the literature.

\vspace*{-0.1cm}
\subsection{Label Augmentation with Superpixels}
\label{subsec:label_aug}
The first approach for propagating point labels to obtain augmented ground truth regions used mean-shift segmentation and edge detection to propose superpixels that were used to extract hand-crafted features for classification with a support vector machine \cite{friedman2013automated}.  Yu \etal~\cite{yu2019fast} proposed a coarse-to-fine segmentation method with label augmentation based on a random selection of pixels from Simple Linear Iterative Clustering (SLIC) \cite{achanta2012slic} superpixel regions. They also proposed a weakly supervised approach to coral segmentation based on Latent Dirichlet Allocation for predicting pseudo-labels for unlabeled image patches \cite{yu2019weakly}. However, both approaches were only evaluated with point labels on a custom dataset of 120 images.  

Alonso \etal~\cite{alonso2018semantic, alonso2019coralseg} designed a multilevel approach that repeatedly uses the Superpixels Extracted by Energy-Driven Sampling (SEEDs) \cite{van2012seeds} algorithm for varying numbers of superpixels in each level.  The levels are combined starting from the largest number of superpixels such that detail from previous levels is preserved \cite{alonso2018semantic, alonso2019coralseg}.  Their approach relies on clustering of color features to produce homogeneous coral regions.  This multilevel approach was improved in \cite{pierce2020reducing} by using an optimized SLIC algorithm and by taking the mode of the class labels across the levels for each pixel in the mask, instead of performing the join operation as in \cite{alonso2018semantic}.

The Superpixel Sampling Network \cite{jampani2018superpixel} was designed as a fully differentiable version of the SLIC algorithm to generate superpixels based on the clustering of features learned by an end-to-end trainable, task-specific architecture.  There have since been numerous other approaches that aim to combine CNNs with the generation of superpixel segments \cite{wang2020end, cai2021revisiting, tu2018learning}.  These models are trained using the semantic segmentation ground truth masks.  

To our knowledge, an approach that uses point labels to inform the creation of the superpixel regions has not been proposed in the literature. In the context of segmentation of underwater imagery, this is an opportunity for improvement of the efficiency and accuracy of point label propagation, which in turn can be used to train a network to perform semantic segmentation of the seafloor.
\begin{figure*}[ht]
\vspace*{0.2cm}
\centering
\includegraphics[width=0.95\textwidth, clip, trim=3cm 11.5cm 3cm 0cm]{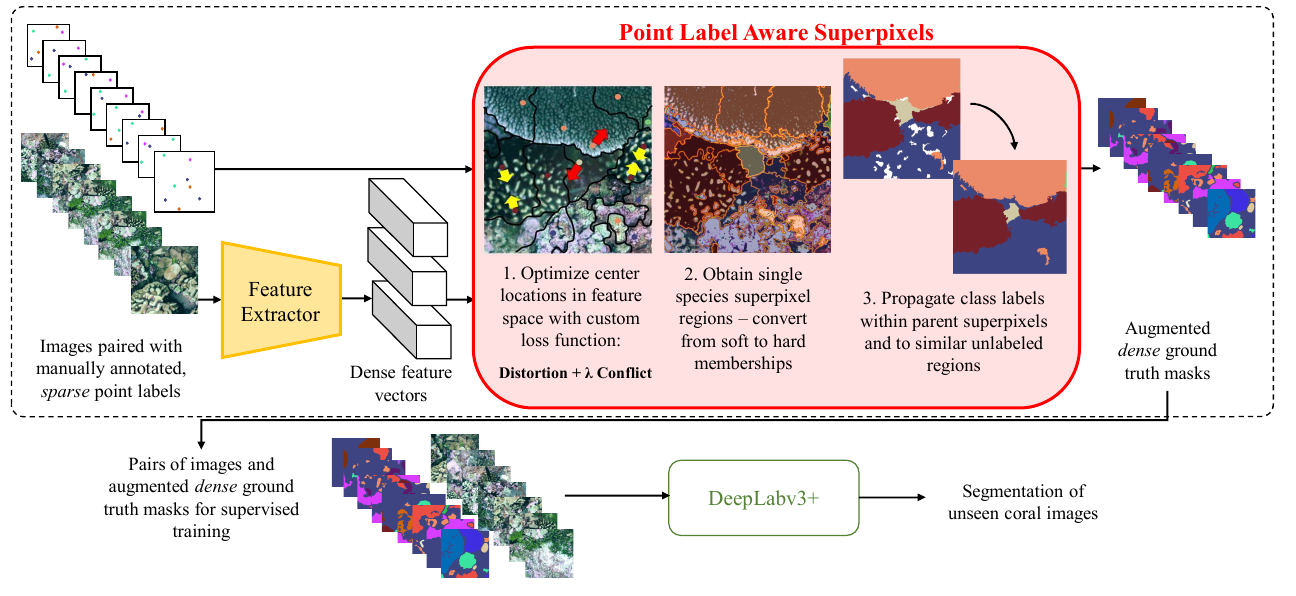}
\vspace*{-0.4cm}
\caption{Proposed Algorithm Schematic. \textbf{Top, Stage One:} Our method takes coral images paired with sparse point labels as input. A feature extractor is used to produce dense feature vectors for the input images, which are then passed into our point label aware superpixel approach. Superpixel centers are optimized using our novel loss function and class labels are propagated within parent superpixels and to similar unlabeled segments. \textbf{Bottom, Stage Two:} The augmented ground truth masks can be used to train a model such as DeepLabv3+ \cite{chen2018encoder} to perform semantic segmentation of unseen coral images.}
\label{fig:pipeline}
\end{figure*}

\section{Method}
\label{sec:method}
Our proposed point-aware superpixel method addresses a clustering problem where the location of each cluster center can be optimized using a loss function (Section~\ref{subsec:methodoverview} and Fig.~\ref{fig:pipeline}). We designed a loss function with two terms to optimize superpixel centers in the embedding space: the distortion loss that creates superpixels based on similarity of pixel features generated by a CNN (Section~\ref{subsec:distort}), and the conflict loss ensures that superpixels do not contain conflicting class labels (Section~\ref{subsec:conflict}). We describe our ensemble method in Section~\ref{subsec:ensemble}.  
It is important to highlight that our novel loss function is not used to train the feature extractor; it instead optimizes superpixel centers in a query image. 

\subsection{Method Overview}
\label{subsec:methodoverview}
Our point label aware approach to superpixels takes a photo-quadrat image of the seafloor and the sparse pixels labeled by an ecologist as input, and generates a dense pseudo ground truth mask for the image. The image and dense ground truth mask pair can then be used for training a supervised semantic segmentation approach. Our method uses our custom loss function to optimize superpixel center locations and maximize the similarity of pixels inside each region, while minimizing the inclusion of conflicting class labels. Specifically, we optimize the cluster centers by minimizing the following loss function,
\begin{equation}
    \label{eq:weightterm}
    \mathcal{L} =  \mathcal{L}_{\mathrm{distortion}} + \lambda \mathcal{L}_{\mathrm{conflict}}
\end{equation}
where $\lambda$ is the scaling factor of the conflict loss term, as determined in Section~\ref{subsec:ablations}.

\vspace*{-0.1cm}
\subsection{Distortion Loss}
\label{subsec:distort}
The distortion loss aims to generate superpixels which neatly conform to coral boundaries.  The distortion loss clusters similar pixels, where each pixel is represented by deep features and its $(x, y)$ location.  

First, for each pixel $p \in P$ and each superpixel $i \in I$, we calculate the Euclidean distances $Q$ between the feature vectors representing the local region centered at the pixel, $T_p$ and the feature vector representing the superpixel, $T_i$ (features are scaled via average L2 norm). The feature vectors are generated using the encoder of the Superpixel Sampling Network (Section \ref{subsec:Implementation}).  We further calculate the Euclidean distance between the $(x, y)$ location (scaled by the number of superpixels along the height/width of the image) of the pixel $X_p$ and the cluster $X_i$,
\begin{equation}
    \label{eq:distances}
    Q_{p,i} =\exp\left({\frac{-{||T_p-T_i||}^2}{2{\sigma_t}^2} + \frac{-{||X_p-X_i||}^2}{2{\sigma_x}^2}}\right),
\end{equation}
where $\sigma_t$ and $\sigma_x$ are the standard deviations for the Gaussian normalization of the distances. These values control the degree of `softness' or `hardness' of the fuzzy memberships and facilitate greater control over the effects of the $(x, y)$ and CNN feature vectors (Section~\ref{subsec:ablations}). 

The fuzzy membership or probability $\mu_{p,i}$ of pixel \textit{p} belonging to the $i$-th superpixel is calculated as the normalized potential
\begin{equation}
    \label{eq:softmemberships}
    \mu_{p,i} = \frac{Q_{p,i}}{\sum_j Q_{p,j}}
\end{equation} for all $|P|$ pixels in the image and $|I|$ superpixels.  

The vector $F_p$ consists of the output of the feature extractor, $T_p$, concatenated with the scaled $(x, y)$ coordinates such that $F_p^T = \left[T_p^T, X_p^T\right]$. The weighted distortion loss is then obtained by weighting the distances between each pixel, $F_p$, and superpixel feature, $S_i^T = \left[T_i^T, X_i^T\right]$, with the probabilities that each pixel belongs to that superpixel,
\begin{equation}
    \label{eq:distortionloss}
    \mathcal{L}_{\mathrm{distortion}}(p) = \sum_{i \in I} ||F_p-S_i||^2 \mu_{p,i}
\end{equation}
which is averaged over the entire image.

We reduce computational time by taking a random subset of pixels per image for calculating the distortion loss.

\subsection{Conflict Loss}
\label{subsec:conflict}
Two conflicting point labels within a superpixel suggest that the superpixel boundary does not effectively separate two objects or classes in the image. In this situation, it is not appropriate to propagate the point label within the superpixel.  

To address this problem, we introduce the conflict loss to generate superpixels which are optimized for the propagation of point labels such that only one class is present in each superpixel.  The conflict loss only considers labeled pixels.  We sum the inner products of the fuzzy memberships for conflicting labeled points,
\begin{equation}
    \label{eq:conflict}
        \mathcal{L}_{\mathrm{conflict}} = \sum_{a,b \in L} \mu_a^T \mu_b \; \mathbb{I}_{l_a \neq l_b}
\end{equation} 
where $L$ is the set of pixels with a point label, $l_p$ is the label at pixel $p$, $\mu_p = \left[\mu_{p,1}, \mu_{p,2}, ..., \mu_{p,|I|}\right]^T$ is the fuzzy membership vector of pixel $p$ with respect to all superpixels, and $\mathbb{I}$ is an indicator function where $\mathbb{I}_B = 1$ if $B$ is true and $0$ otherwise.

The conflict loss is large when many pixels have conflicting labels and similar fuzzy memberships; and small if all conflicting point labels have dissimilar fuzzy memberships, indicating the points are members of distinct segments.  We also average the conflict loss across the labeled pixels with conflicting class labels.

\begin{figure}
    \centering
    \includegraphics[width=0.95\columnwidth, clip, trim=0cm 6.5cm 0cm 4cm]{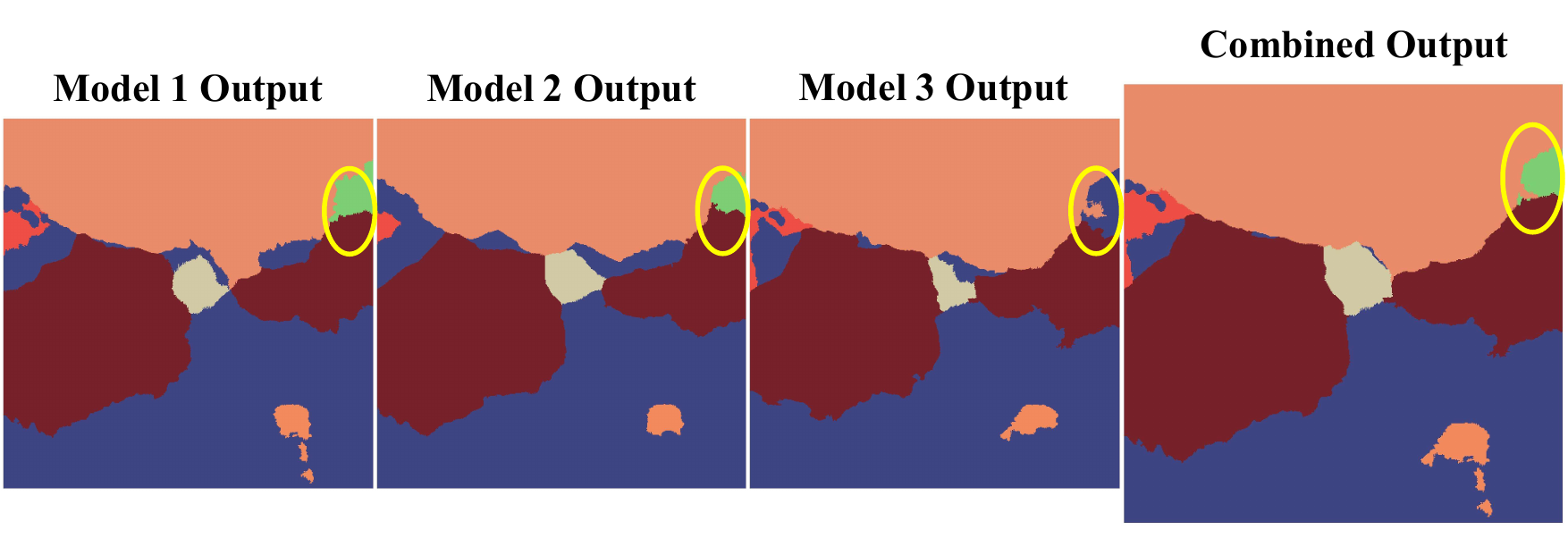}
    \vspace{-0.5cm}
    \caption{The ensemble method combines the output from three of our point label aware superpixel models.  In the output from the third model, the superpixel segments have resulted in a missed region of Corallimorpharia (depicted in light green).  The ensemble method ensures the output mask preserves the correct propagated pixels of this class from the other two outputs, resulting in a more robust augmented ground truth mask.}
    \vspace{-0.2cm}
    \label{fig:ensemble}
\end{figure}

\subsection{Ensemble Method}
\label{subsec:ensemble}
Up to this point, we performed point label propagation by inspecting each superpixel and checking if there is a point label present in the region.  If we have superpixels without any point labels inside, we found the most similar labeled superpixel in the image and propagated that class (see Section~\ref{subsec:Implementation} for details). Therefore, our augmented ground truth will be more accurate if the number of unlabeled superpixels is reduced. 

To this end, we design an ensemble method to generate our superpixel regions three times, using three models with distinct hyperparameter choices to encourage three different superpixel decompositions.  In this way, we maximize the likelihood that superpixel regions contain point labels and reduce the uncertainty involved with propagating the class from the most similar labeled region. We combine the three propagated ground truth masks by taking the mode at each pixel.  We ensure information is preserved by ignoring the `unknown' class, except in the case that all classifiers yield this label.  Fig.~\ref{fig:ensemble} demonstrates that the ensemble improves the robustness of our approach.

\begin{table*}
\vspace*{0.2cm}
\caption{Stage One Results: Performance of Label Propagation Approaches (Refer to Section~\ref{subsec:evaluationmetrics} for Metric Definitions)}
\centering

\begin{tabular}{@{}lcccc}
\toprule

\textbf{Method}        & {\textbf{PA}} & {\textbf{mPA}} & {\textbf{mIoU}}  & {\textbf{Time per Image (s)}}    \\
\midrule
CoralSeg Multilevel Algorithm using SLIC and 15 Levels \cite{alonso2019coralseg} & 88.94 & \textit{87.00} & 76.96 & 5.72 \\
Fast Multilevel Algorithm \cite{pierce2020reducing} & 86.55 & 83.70 & 79.75 & 3.07 \\
Superpixel Sampling Network \cite{jampani2018superpixel} & 88.06 & 83.52 & 79.19 & 1.46 \\
\midrule

Point Label Aware Superpixels (Ours) without feature extractor -- raw LAB values & 64.14 & 52.39 & 47.27 & \textbf{0.79} \\
Point Label Aware Superpixels (Ours) with ResNet-18 \cite{he2016deep} pretrained on ImageNet & 87.21 & 83.66 & 81.30 & \textit{0.86} \\
Point Label Aware Superpixels (Ours) with ResNet-18 \cite{he2016deep} pretrained on ImageNet and & \multirow{2}{*}{88.22} & \multirow{2}{*}{84.88} & \multirow{2}{*}{82.29} & \multirow{2}{*}{\textit{0.86}} \\
fine-tuned on UCSD & & & & \\
Point Label Aware Superpixels (Ours) with SSN \cite{jampani2018superpixel} encoder with random initialization & 87.85 & 83.07 & 81.89 & 1.37 \\
\midrule
Point Label Aware Superpixels (Ours) with SSN \cite{jampani2018superpixel} encoder trained on UCSD \textit{- Single Classifier} & \textit{89.61} & 86.75 & \textit{83.96} & 1.37\\
Point Label Aware Superpixels (Ours) with SSN \cite{jampani2018superpixel} encoder trained on UCSD \textit{- Ensemble} & \textbf{92.56} & \textbf{89.47} & \textbf{85.31} & 3.34 \\

\bottomrule
\end{tabular}\\
\label{tab:augmentation}
\end{table*}
\section{Experimental Setup}
\label{sec:experimentalsetup}
In this section, we briefly discuss implementation details (Section~\ref{subsec:Implementation}), evaluation datasets (Section~\ref{subsec:datasets}), and employed evaluation metrics (Section~\ref{subsec:evaluationmetrics}).

\begin{table*}
\caption{Stage Two Results: Comparison of Approaches Trained with the Augmented Ground Truth\\on the UCSD Mosaics and Eilat Datasets (Refer to Section~\ref{subsec:evaluationmetrics} for Metric Definitions)}
\centering
\begin{tabular}{@{}c@{\hspace{0.5cm}}ccc@{\hspace{0.5cm}}ccc}
\toprule
        & \multicolumn{3}{c}{\textbf{USCD Mosaics}} & \multicolumn{3}{c}{\textbf{Eilat}}  \\
\cmidrule(lr){2-4}
\cmidrule(lr){5-7}
\textbf{Method} & PA & mPA & mIoU & PA & mPA & mIoU \\
\midrule
SegNet trained with Single-level Superpixels \cite{alonso2018semantic} & -- & -- & -- & 81.23 & 41.97 & 28.14 \\
DeepLabv3+ trained with CoralSeg Multilevel Algorithm \cite{alonso2019coralseg} & 86.11 & 59.90 & 49.16 & 84.80 & 54.65 & 44.01 \\
DeepLabv3+ trained with Superpixel Sampling Network \cite{jampani2018superpixel} & 88.42 & 66.68 & 56.74 & 86.46 & 60.45 & 52.60 \\
\midrule
DeepLabv3+ trained with Point Label Aware Superpixels (Ours) & \textbf{89.02} & \textbf{67.97} & \textbf{58.81} & \textbf{88.99} & \textbf{65.87} & \textbf{58.33} \\
\bottomrule
\end{tabular}
\label{tab:deeplab}
\end{table*}

\subsection{Implementation}
\label{subsec:Implementation}
All experiments are conducted with an NVIDIA GeForce RTX 2080, and inference times are with respect to this GPU. In the following, we discuss the hyperparameters and implementation details for the two stages (\ie, dense label generation and DeepLab training) separately.

\subsubsection{Stage One: Dense label generation}
We implemented our superpixel method using PyTorch \cite{paszke2019pytorch}. We determined that suitable values for the Gaussian normalization terms are $\sigma_t=0.5534$ and $\sigma_x=0.631$ (as shown in the ablation study in Fig.~\ref{fig:sigma}).  We use $\lambda=1140$ to weight the conflict loss term (Fig.~\ref{fig:lambda}), and we find that only 3000 pixels are needed for calculating the distortion loss (as opposed to using all image pixels, see Eqs.~\ref{eq:softmemberships} and \ref{eq:distortionloss}), significantly reducing the computation time.  We use 100 superpixels per image. We use the encoder architecture from the Superpixel Sampling Network (SSN) \cite{jampani2018superpixel} to extract dense feature vectors $T_p$ representing the region around each pixel in the image. 

We propagate point labels to their parent superpixels: using the hard memberships, we inspect each superpixel and see if there is a point label inside the region. If so, we copy the class of the label to all other pixels within that superpixel.  If multiple point labels and differing classes are present in the superpixel, we take the majority vote. For superpixels that do not contain any point labels, we find the labeled superpixel with the most similar feature vector and propagate the class associated to that superpixel.  We also apply this propagation method when evaluating the original Superpixel Sampling Network to ensure a fair comparison. 

For our ensemble method (Section~\ref{subsec:ensemble}), we choose hyperparameter values within our recommended ranges (as established in the ablation studies in Figs.~\ref{fig:lambda} and~\ref{fig:sigma}) and vary the values slightly to encourage some variation in the generation of the superpixels\footnote{Based on the recommended ranges identified in Figs.~4 and 5, for the first model we use $\sigma_t=0.5539$, $\sigma_x=0.5597$ and $\lambda=1500$; for the second model we take $\sigma_t=0.846$, $\sigma_x=0.5309$ and $\lambda=1590$; and $\sigma_t=0.553$, $\sigma_x=0.631$ and $\lambda=1140$ for the third.}.

\subsubsection{Stage Two: DeepLab training}
To evaluate the efficacy of the augmented ground truth masks for training a model to perform semantic segmentation, we use Tensorflow to train DeepLabv3+ \cite{chen2018encoder}.  We use data augmentation consisting of random horizontal and vertical flipping, gain ($0.8 - 1.2$) and gamma ($0.8 - 1.2$) and train for 500 epochs using the Adam optimizer with a learning rate of 0.001.  Consistent with the results reported in \cite{alonso2019coralseg}, results are reported on the test dataset for the best epoch during training.
 
\subsection{Datasets}
\label{subsec:datasets}
UCSD Mosaics is the only publicly available, multi-species coral dataset that provides dense ground truth masks \cite{edwards2017large, alonso2019coralseg}.  We use the version of the dataset provided by \cite{alonso2019coralseg} to ensure fair comparison.  Each image is 512 by 512 pixels, yielding 262,144 labeled pixels per image. We take a random $\approx$0.1\% of the dense labels (300 labels) for demonstrating our point label aware superpixel approach and perform evaluation using the dense labels.  For consistency with \cite{alonso2019coralseg}, we ignore the `background' class during evaluation.

We also evaluate our method on the Eilat Fluorescence dataset \cite{beijbom2016improving}, which consists of 142 training images and 70 test images labeled using 200 sparse point labels arranged as a grid in the center of each image. The points are labeled into ten classes.  The images were originally 3K by 5K pixels, with RGB and wide-band fluorescence information for each pixel.  We do not use the fluorescence data in our approach and we therefore do not compare our method to approaches which leverage this information.  Following \cite{alonso2019coralseg}, we downsize the images to 1123 by 748 pixels. We use our point label aware superpixel approach as for the UCSD Mosaics dataset, however for unlabeled superpixels we set the class as `substrate' instead of choosing the most similar labeled segment, due to the large apparatus present in each of the images. Although this dataset does not have dense ground truth masks, it has been used to evaluate prior approaches for coral classification and segmentation \cite{alonso2019coralseg, xu2019coral, nadeem2019deep}.

\subsection{Evaluation Metrics}
\label{subsec:evaluationmetrics}
We use standard metrics for semantic segmentation as commonly used in the literature \cite{alonso2019coralseg,garcia2018survey}: pixel accuracy (PA), \ie the sum of correctly classified pixels divided by the predicted pixels, the mean pixel accuracy (mPA), which is the pixel accuracy averaged over the classes, and the mean intersection over union (mIoU), which is the average of the per-class IoU scores (for all three metrics, a higher score reflects better performance).  For consistency with \cite{alonso2019coralseg}, we do not include pixels which are `unlabeled' or `unidentified' when calculating the mIoU score. These metrics are used for evaluation of both Stage One and Stage Two performance. 

\section{Results}
We first compare our method to the state-of-the-art in Section~\ref{subsec:sota} and then provide ablation studies with respect to the loss term weighting, feature extraction method, and fuzzy membership degree in 
Section~\ref{subsec:ablations}.

\begin{figure}
    \centering
    \includegraphics[width=.9\columnwidth]{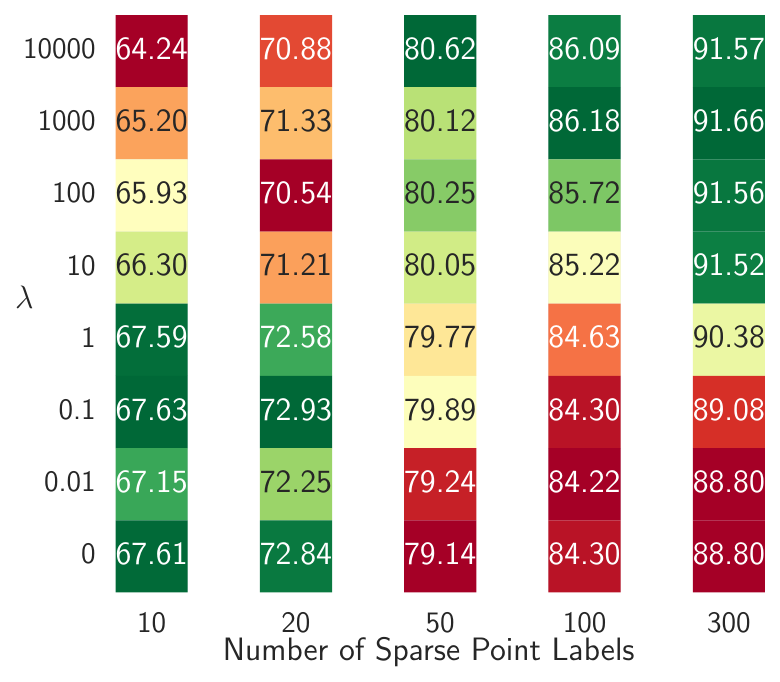}
    \vspace{-0.3cm}
    \caption{Pixel accuracy of label propagation on UCSD Mosaics. For different quantities of sparse point labels, the relative weighting of the two loss terms can be tuned: $\mathcal{L} =  \mathcal{L}_{\mathrm{distortion}} + \lambda \mathcal{L}_{\mathrm{conflict}}$. For smaller quantities of point labels, our method performs best for $\lambda < 1$. When a larger number of labeled pixels is available \eg~300, our method performs optimally for $\lambda \geq 10$.}
    \label{fig:lambda}
    \vspace*{-0.2cm}
\end{figure}

\subsection{Comparison to State-of-the-art Methods}
\label{subsec:sota}
We compare the performance of our novel method to state-of-the-art approaches, namely CoralSeg~\cite{alonso2019coralseg} and the Fast Multilevel Algorithm~\cite{pierce2020reducing}. We also introduce the Superpixel Sampling Network~\cite{jampani2018superpixel} as a competitive baseline for the task of generating augmented ground truth from sparse point labels. 

CoralSeg \cite{alonso2019coralseg} produces augmented ground truth masks by generating superpixels with SLIC \cite{achanta2012slic} for 15 levels, where the first level contains 10 times the number of point labels (3000) and the final level contains one tenth of the point labels (30).  Levels are combined using a join operation which preserves fine-grained details from the initial levels.  The Fast Multilevel Algorithm \cite{pierce2020reducing} improves the speed of CoralSeg by replacing the join operation with a mode calculation for each pixel, and by using a faster version of SLIC. The Superpixel Sampling Network is a fully differentiable, end-to-end trainable version of SLIC which produces task-specific superpixels \cite{jampani2018superpixel}.

\subsubsection{Stage One: Dense label generation}
As shown in Table~\ref{tab:augmentation}, our point label aware approach to superpixels outperforms all baseline approaches, for augmentation of ground truth for coral images in terms of pixel accuracy and mIoU. The relative increase in mIoU is 8.35\%, 5.56\% and 6.12\% when compared to CoralSeg, the Fast Multilevel Algorithm, and the Superpixel Sampling Network, respectively.  Our method also significantly improves computation time per image, to an average of 1.37s for a single classifier or 3.34s for our ensemble method (compared to the second-best performing method, CoralSeg, which takes 5.72s per image). Fig.~\ref{fig:qualitativeresults} presents some qualitative examples demonstrating that our approach generates detailed, single species segments that closely conform to complex coral boundaries.  

\subsubsection{Stage Two: DeepLab training}
Table~\ref{tab:deeplab} demonstrates that our improved generation of augmented ground truth labels also improves the performance of semantic segmentation models, on both the UCSD Mosaics and Eilat datasets. These results are obtained by training DeepLabv3+ \cite{chen2018encoder} on our augmented ground truth. We outperform CoralSeg and the Superpixel Sampling Network on all three metrics; specifically for mIoU by 9.65\% and 2.07\% respectively on UCSD Mosaics and by 14.54\% and 5.95\% on Eilat.

\begin{figure}
    \centering
    \includegraphics[width=.9\columnwidth]{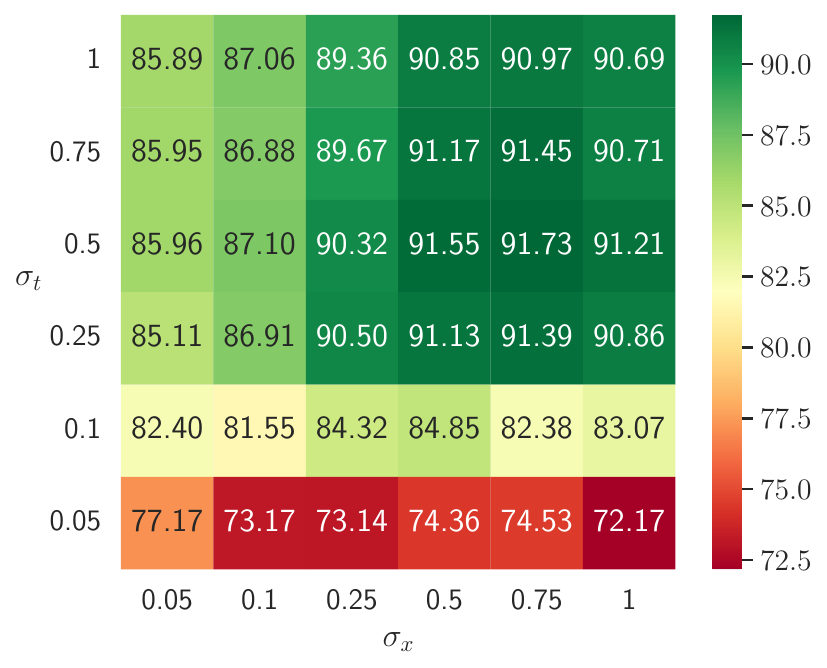}
    \vspace{-0.3cm}
    \caption{Pixel accuracy of label propagation on the UCSD Mosaics dataset. Our point label aware loss function exhibits the best performance in terms of pixel accuracy when values of $\sigma_x$ and $\sigma_t$, which control the softness of the fuzzy memberships based on the $(x, y)$ coordinates and feature vectors respectively, are in the range 0.5-1.}
    \vspace*{-0.2cm}
    \label{fig:sigma}
\end{figure}

\subsection{Ablation Study}
\label{subsec:ablations}

\subsubsection{Weighting the Loss Terms \texorpdfstring{($\lambda$)}{}}
\label{subsubsec:lambda}
We first evaluate the impact of the weight $\lambda$ in Eq.~\ref{eq:weightterm}. The higher $\lambda$, the higher the impact of the conflict loss (and thus the lower the impact of the distortion loss). Fig.~\ref{fig:lambda} shows that if there are a sufficiently large number of random point labels in our image, the conflict loss term plays a greater role in generating superpixels.  If there are very few labels, \eg, 10 points, the distortion loss term plays a more significant role.

\newcommand{\scaleMaskSets}{0.15\textwidth}
\begin{figure*}
    \vspace*{0.2cm}
    \centering
    \begin{tabular}{ccccc}
    \small Query Image & \small Ground Truth & \small CoralSeg~\cite{alonso2019coralseg} & \small SSN~\cite{jampani2018superpixel} & \small Ours \\
    
    \includegraphics[width=\scaleMaskSets]{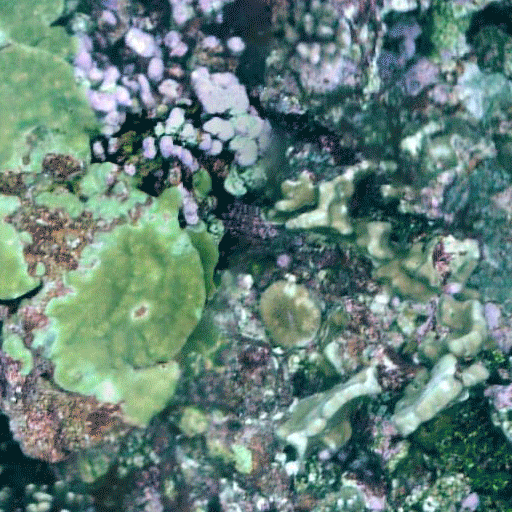} &
    \includegraphics[width=\scaleMaskSets]{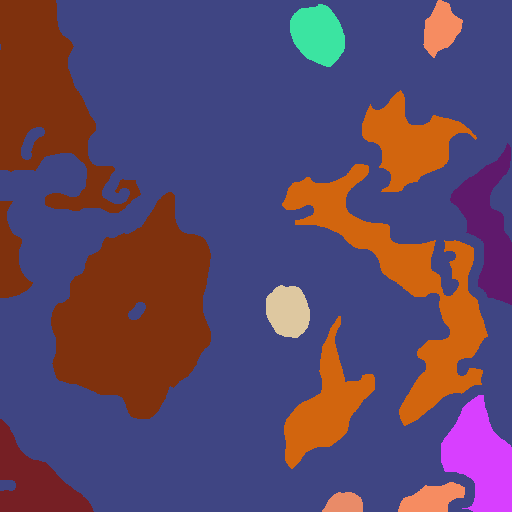} &
    \includegraphics[width=\scaleMaskSets]{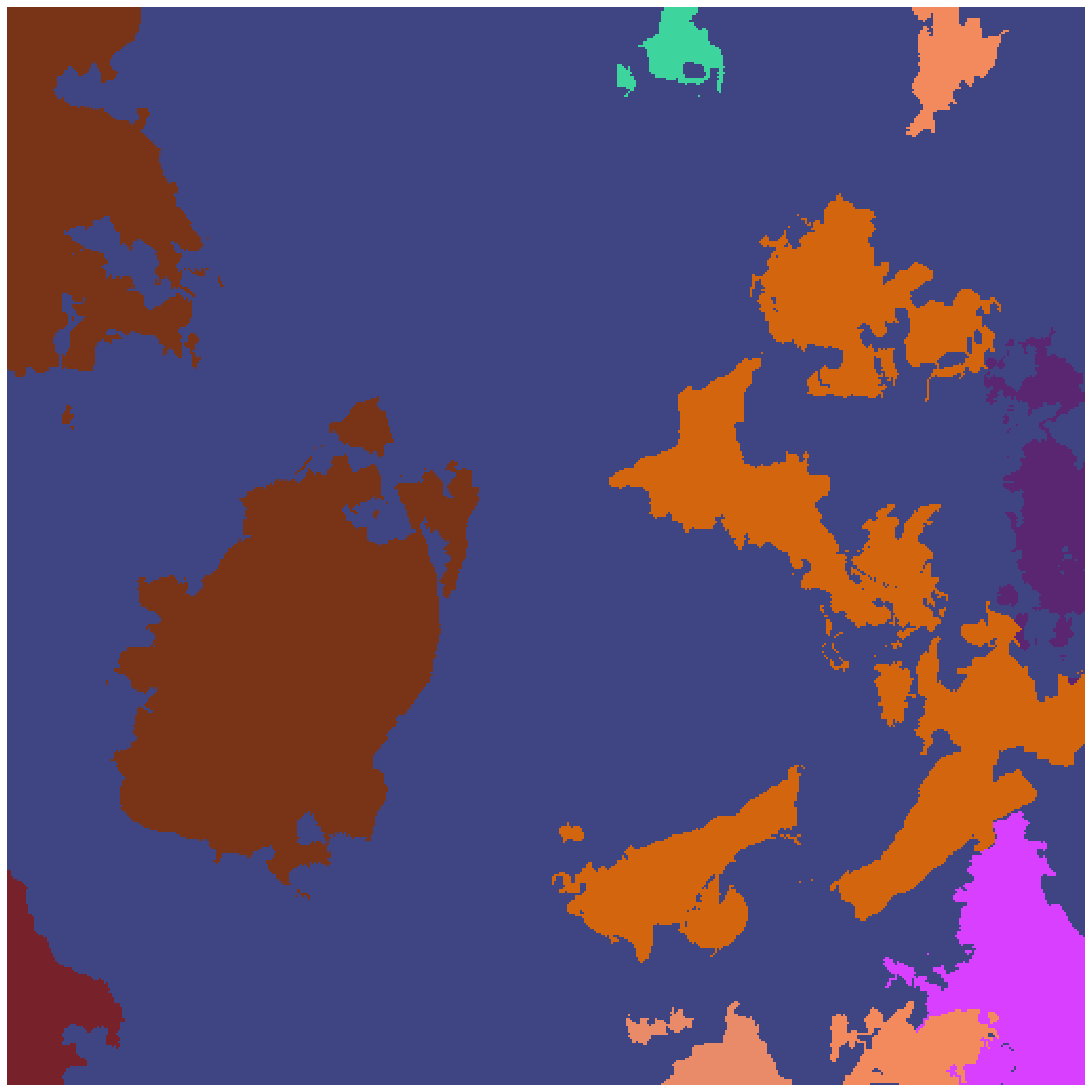} &
    \includegraphics[width=\scaleMaskSets]{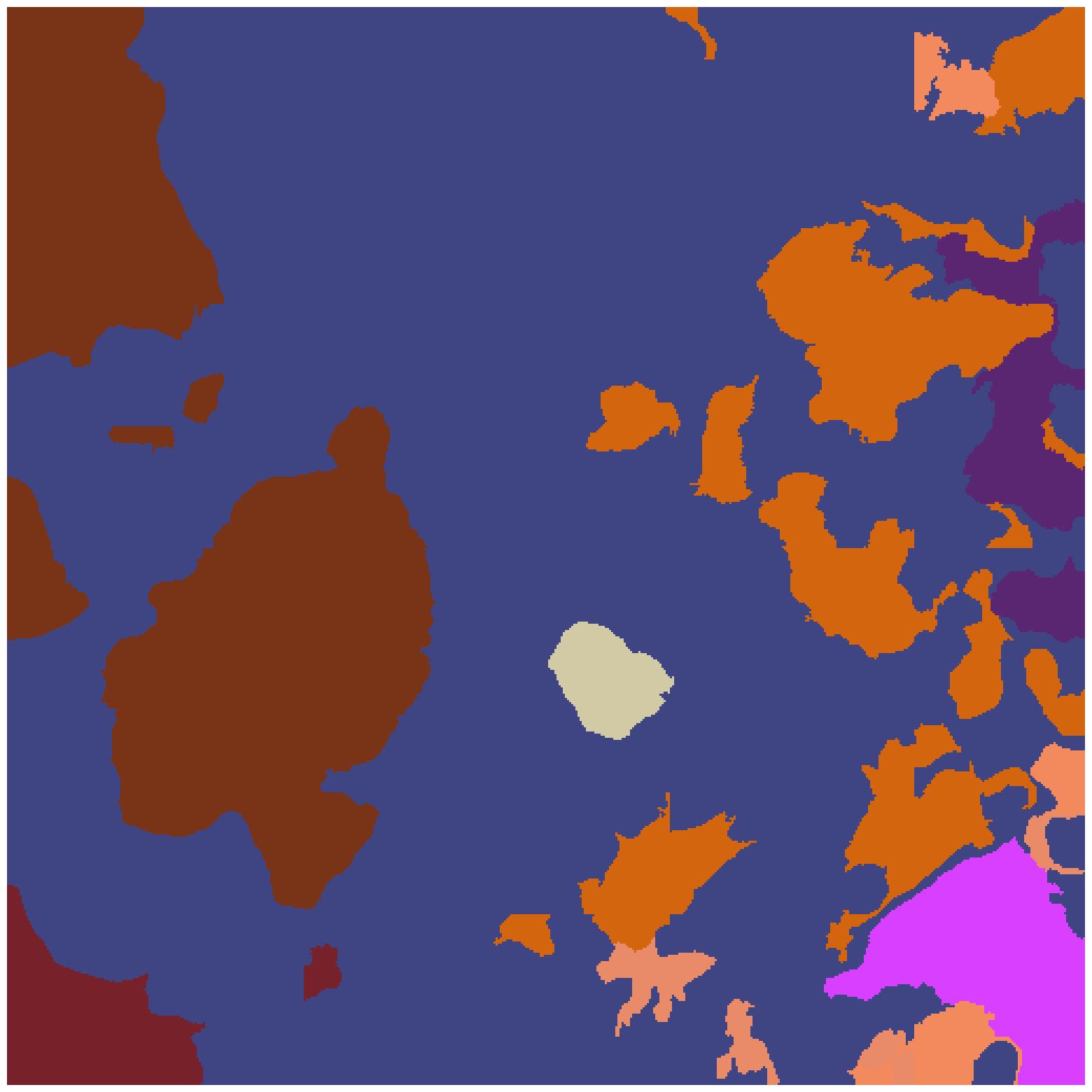} &
    \includegraphics[width=\scaleMaskSets]{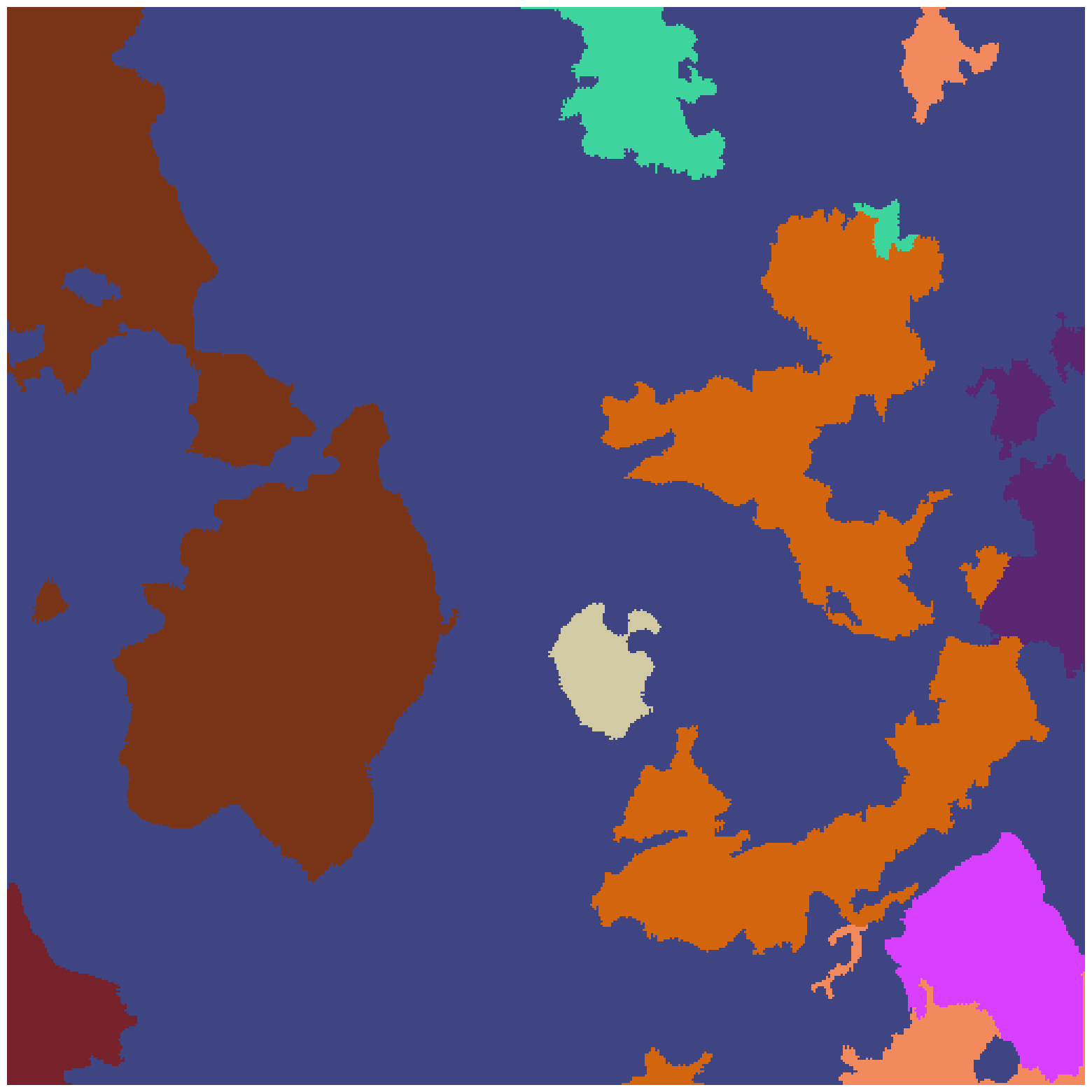} \\    
    
    \includegraphics[width=\scaleMaskSets]{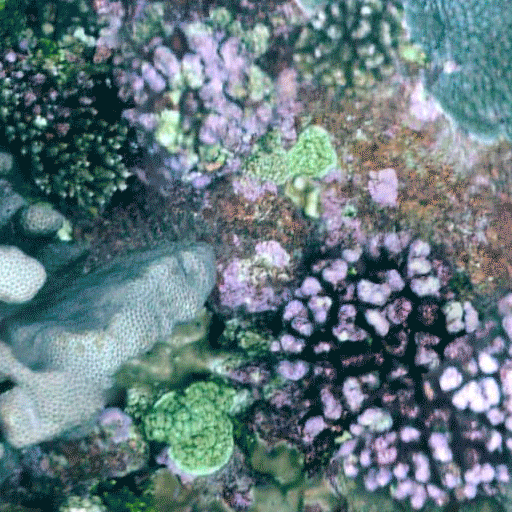} &
    \includegraphics[width=\scaleMaskSets]{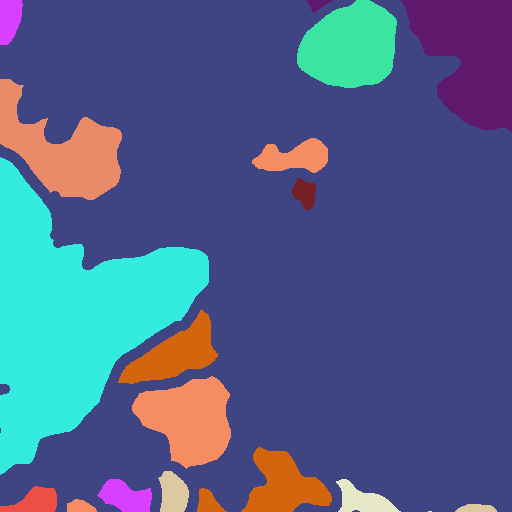} &
    \includegraphics[width=\scaleMaskSets]{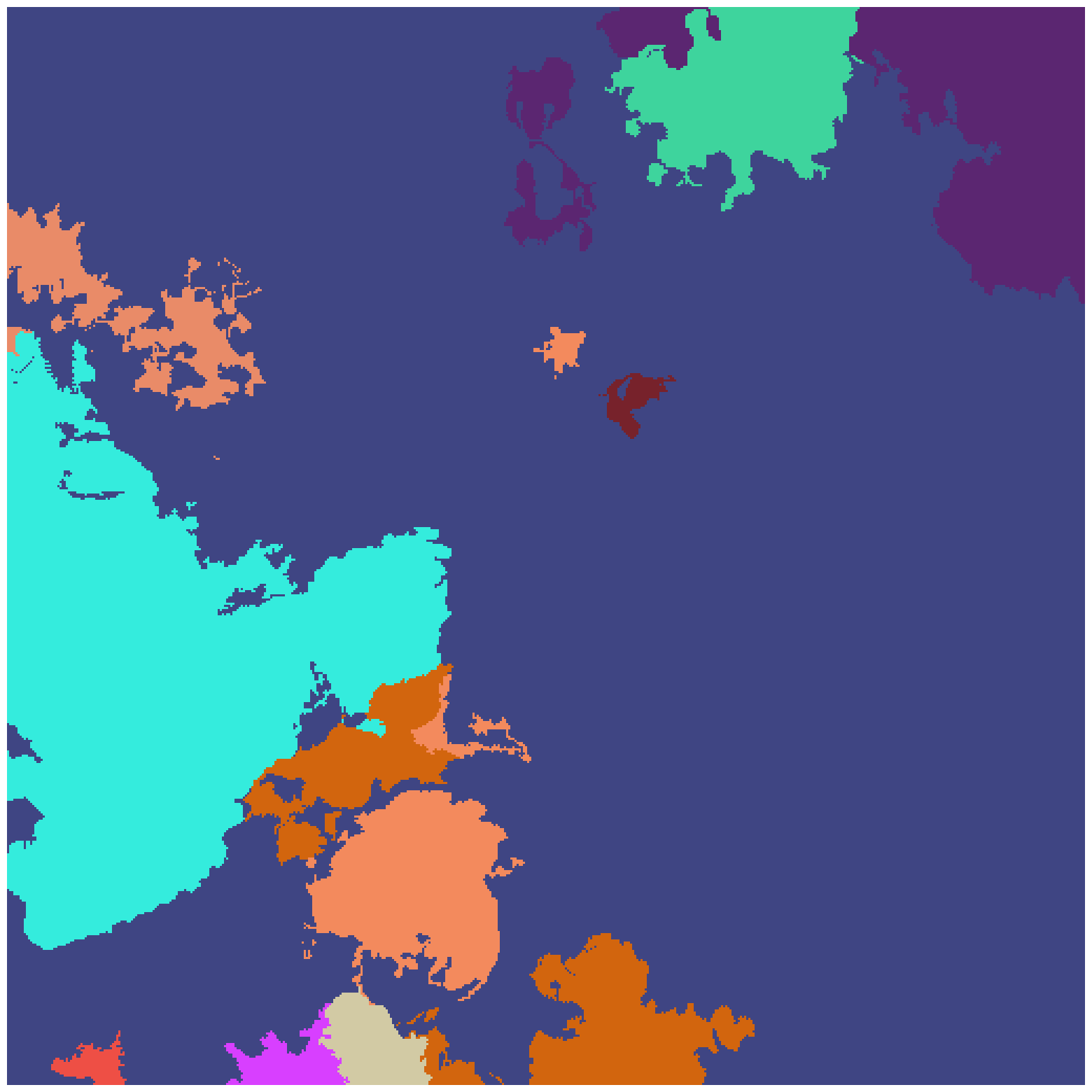} &
    \includegraphics[width=\scaleMaskSets]{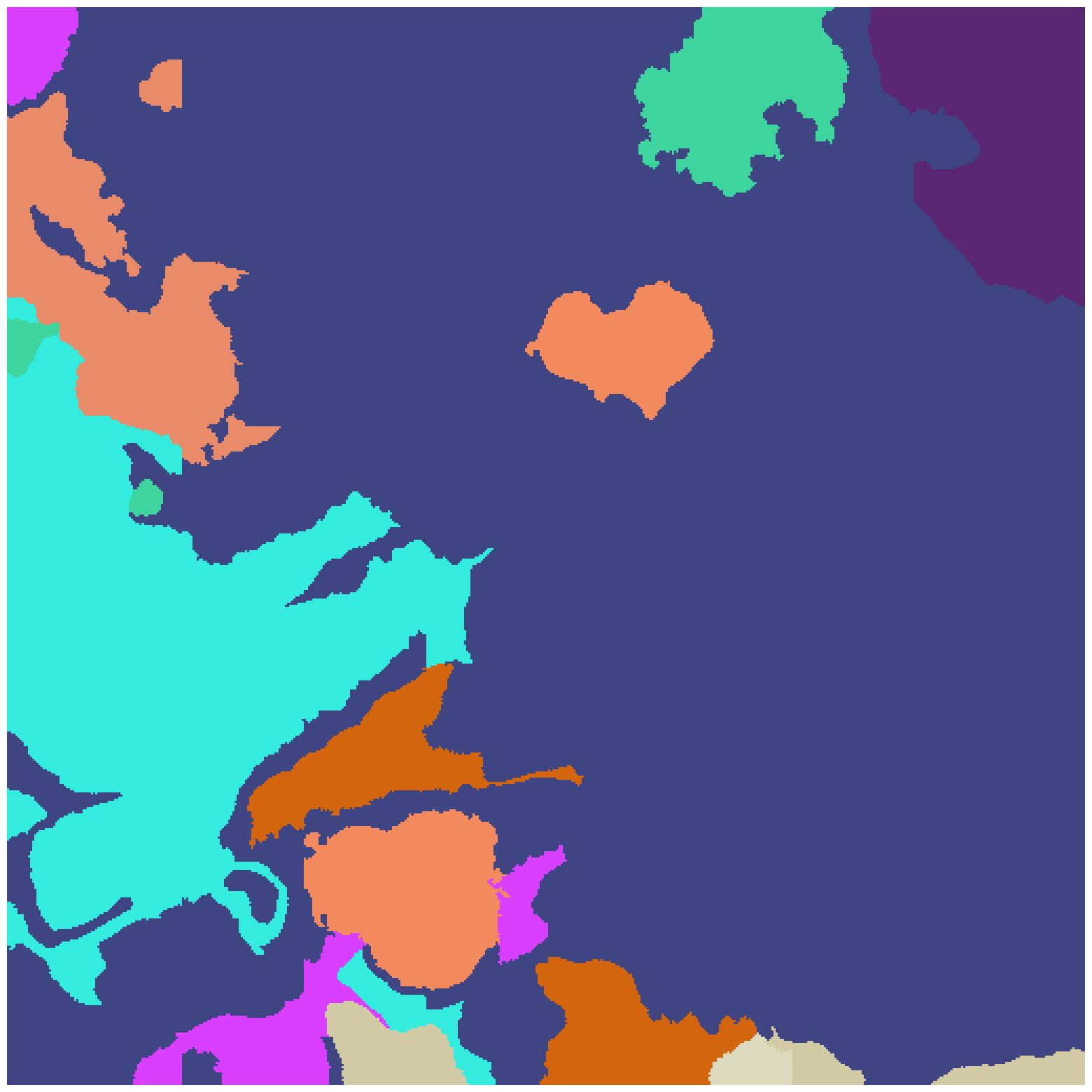} &
    \includegraphics[width=\scaleMaskSets]{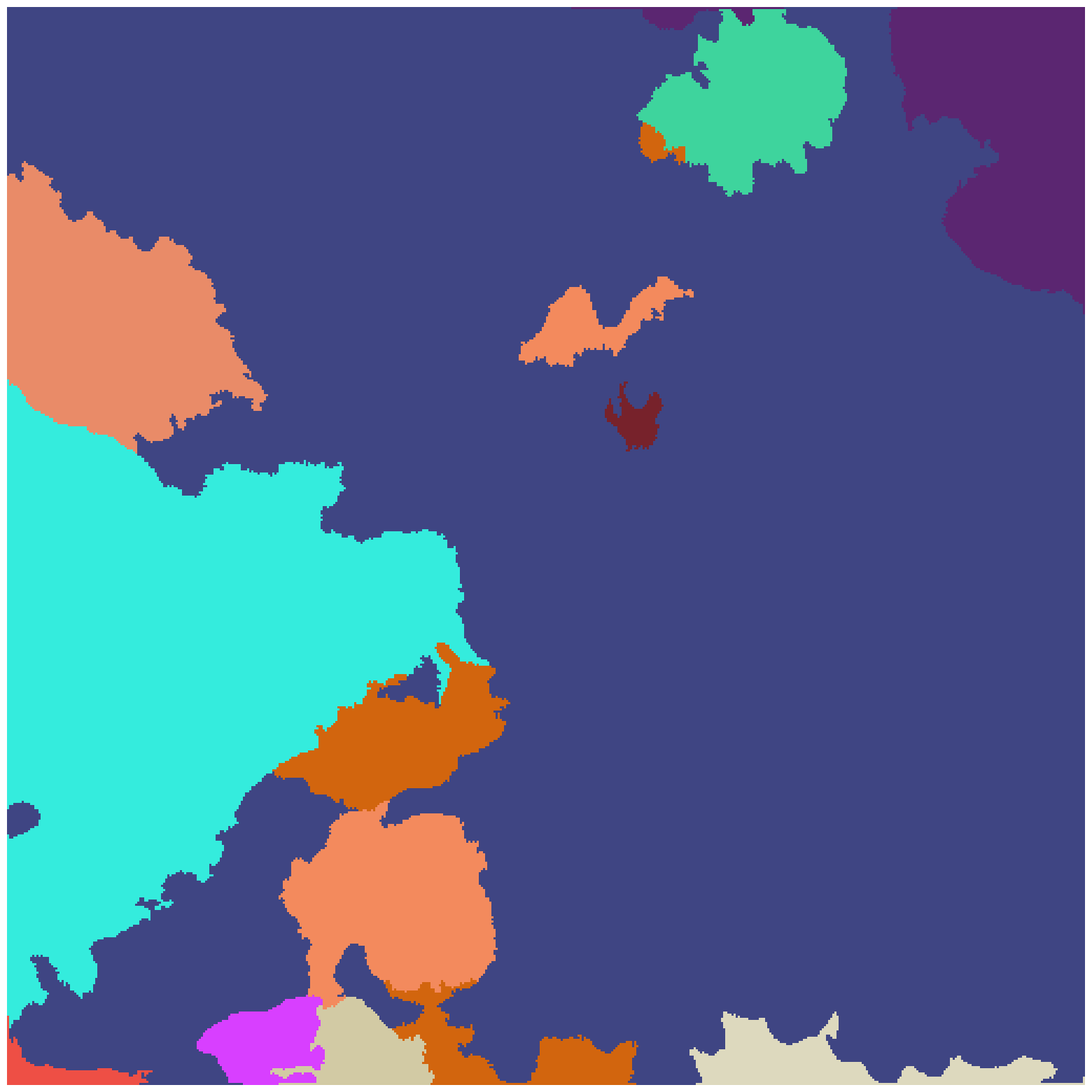} \\
    
    \includegraphics[width=\scaleMaskSets]{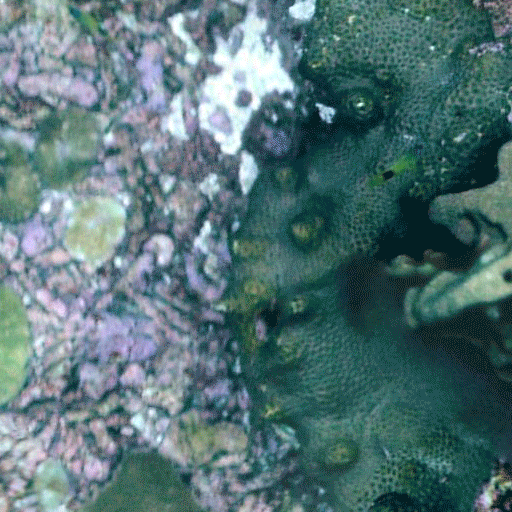} &
    \includegraphics[width=\scaleMaskSets]{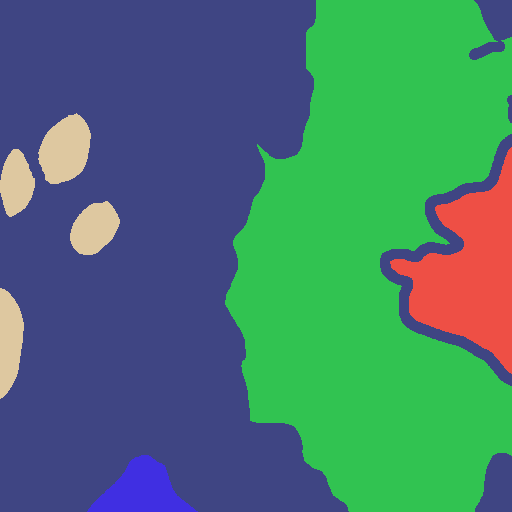} &
    \includegraphics[width=\scaleMaskSets]{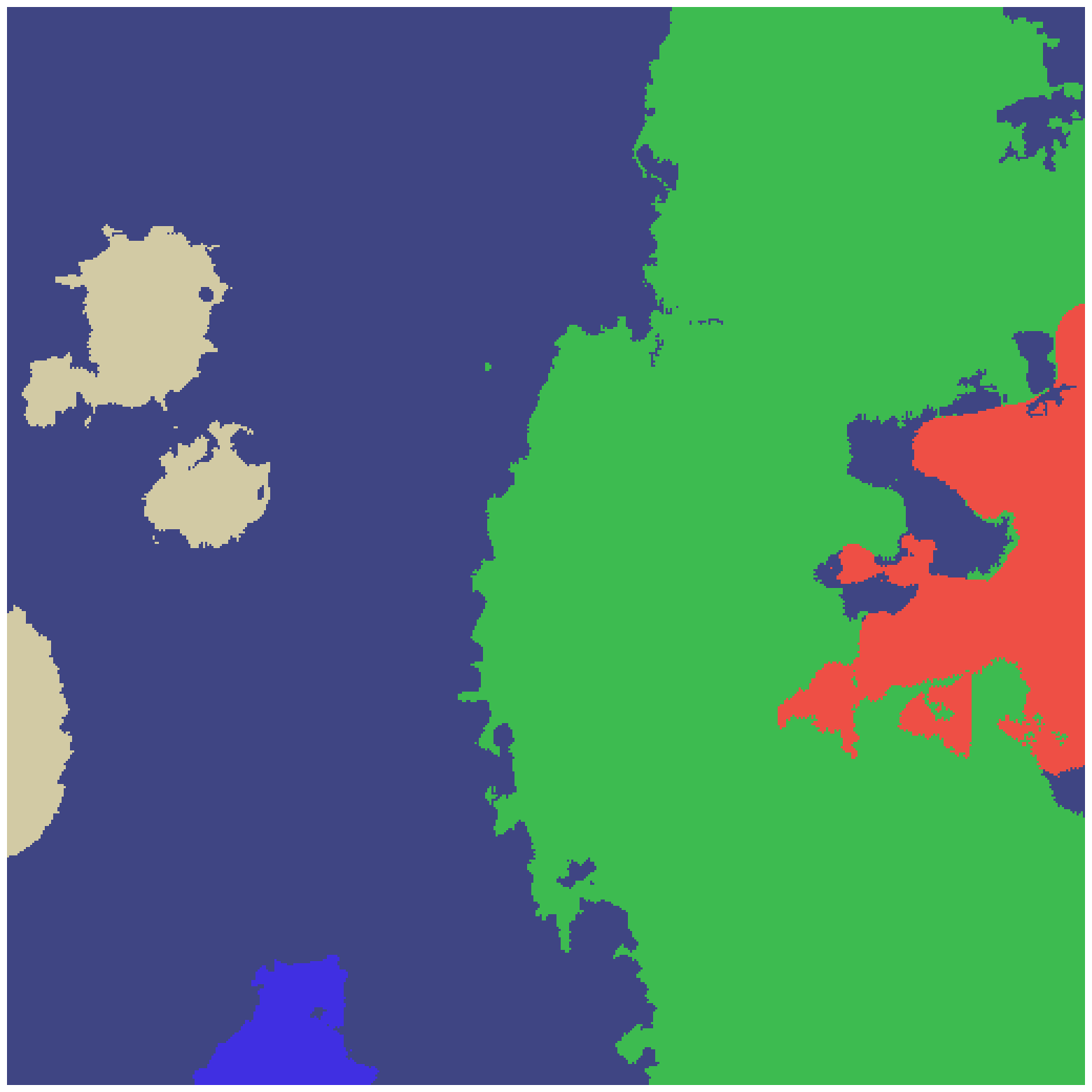} &
    \includegraphics[width=\scaleMaskSets]{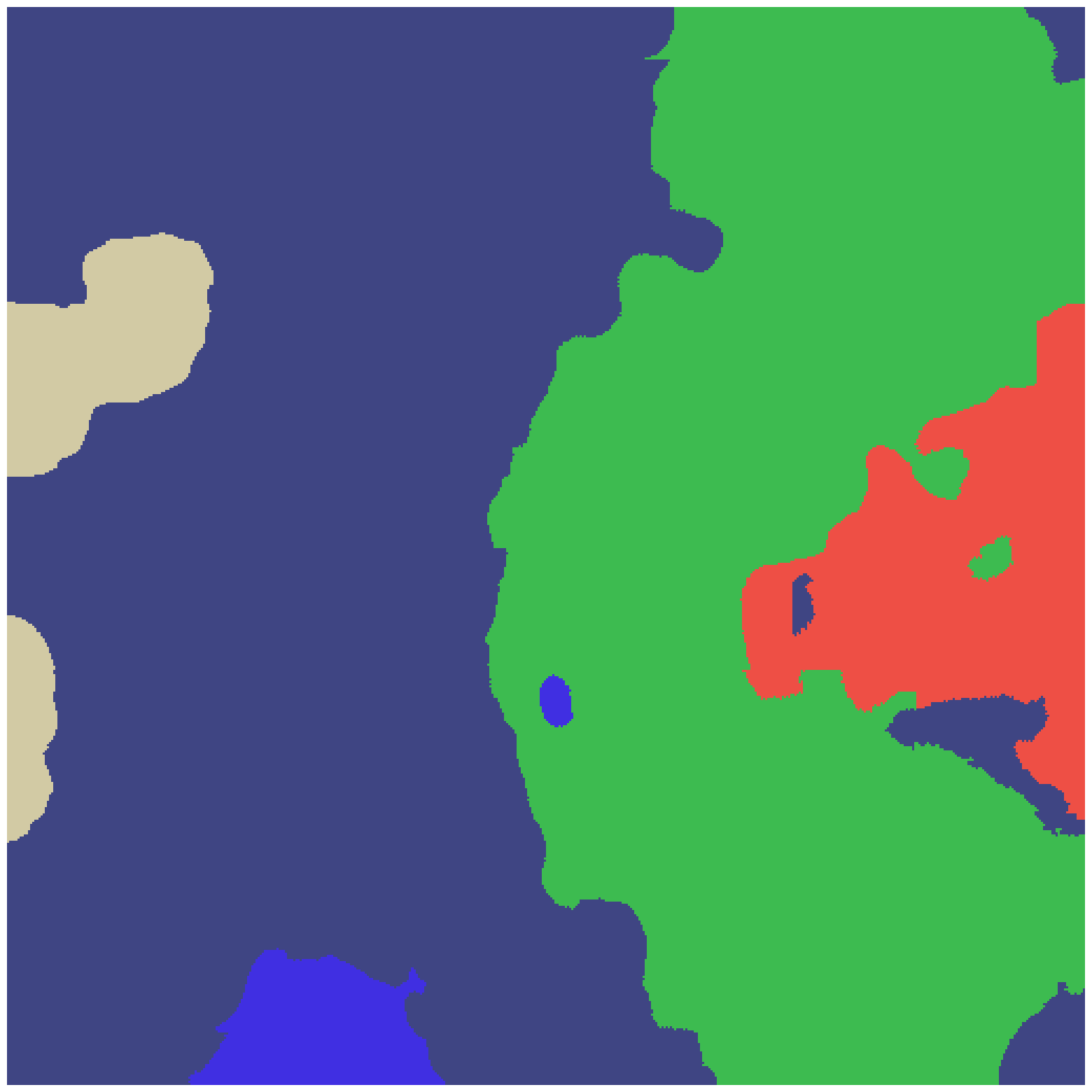} &
    \includegraphics[width=\scaleMaskSets]{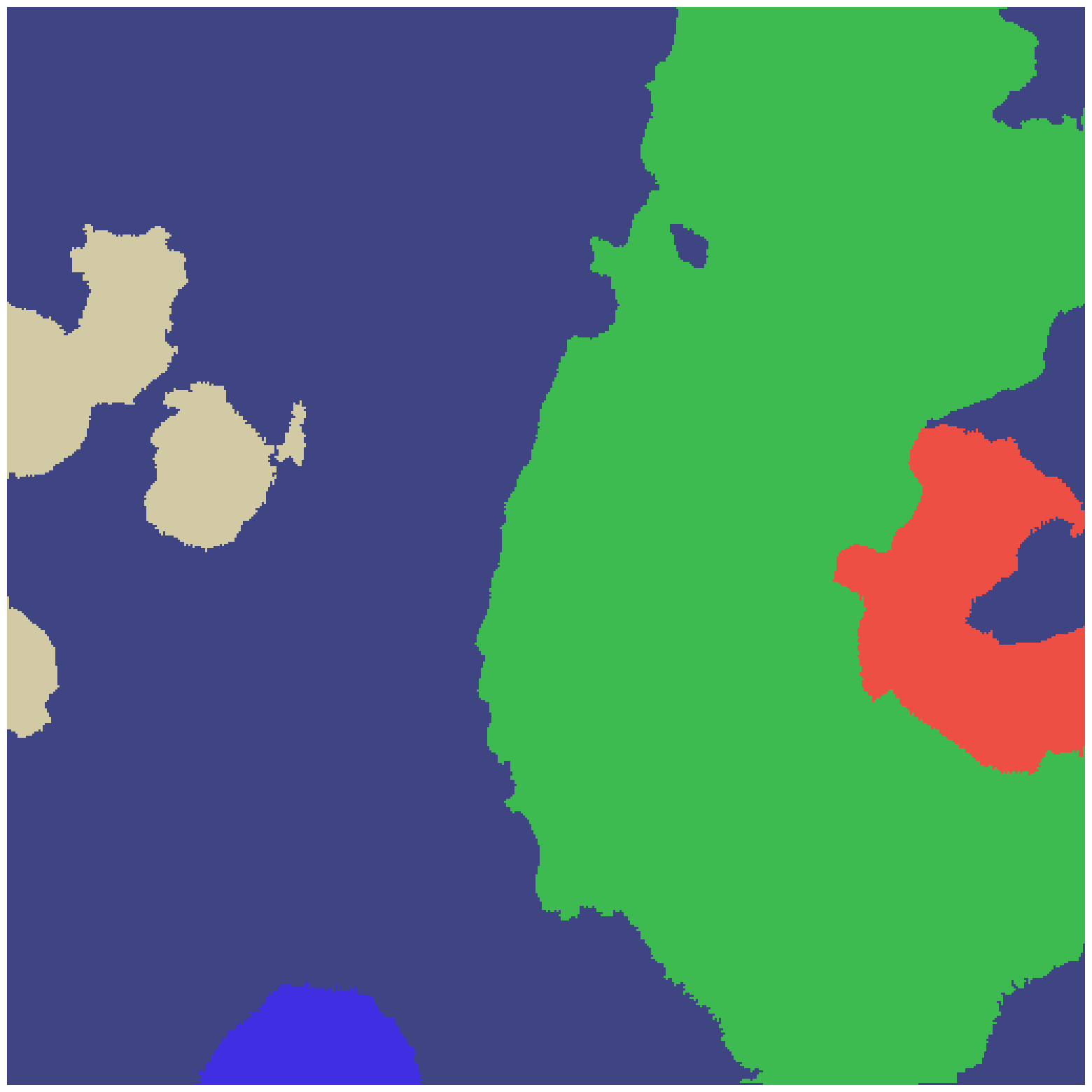} \\
\end{tabular}
    \caption{Qualitative comparison between CoralSeg multilevel \cite{alonso2019coralseg}, Superpixel Sampling Network (SSN) \cite{jampani2018superpixel} and our point label aware superpixel approach.  The top row shows that only our approach effectively captures both the pale green coral segment and the beige segment in the center of the image.}
    \vspace*{-0.3cm}
    \label{fig:qualitativeresults}
\end{figure*}

\subsubsection{Feature Extraction}
\label{subsubsec:features}
We evaluated two different feature extractors -- the fully convolutional encoder from the Superpixel Sampling Network (SSN) \cite{jampani2018superpixel} and a ResNet-18 \cite{he2016deep} encoder pre-trained on ImageNet. We train both models on the UCSD Mosaics dataset and take the encoders for use as the feature extractor in our superpixel approach (Fig.~\ref{fig:pipeline}). For ResNet-18, we extract features after the third convolutional block and bilinearly upsample the output to obtain a dense feature vector for each pixel in our original image.  We also evaluated both encoders as feature extractors without any training on the UCSD Mosaics dense labels. Table~\ref{tab:augmentation} shows that our superpixel approach is close to the same performance in this case. 

Therefore, our superpixel algorithm (that optimizes the superpixel centers) is not sensitive to the pixel features, and it is thus possible to leverage our point label aware approach if no densely labeled images are available for fine-tuning a feature extractor. In other words, our superpixel method can be used to generate augmented ground truth for previously unseen locations and species, given images accompanied by sparse point labels.

\subsubsection{Sigma Values}
\label{subsubsec:sigma}
Fig.~\ref{fig:sigma} shows that our point label aware loss function performs best when values for $\sigma_x$ and $\sigma_t$ are in the range 0.5-1 (remember that $\sigma_x$ controls the softness of the fuzzy memberships based on the scaled $(x, y)$ locations, and $\sigma_t$ controls the softness based on the CNN feature vectors; see Eq.~\ref{eq:distances}).  Performance can be optimized by having a slightly larger value for $\sigma_x$, allowing location-based superpixel memberships to be softer and therefore conform to coral textures more closely.  Decreasing $\sigma_x$ is equivalent to weighting the $(x, y)$ component more highly, resulting in more regular and compact superpixels. For application to coral images, we desire superpixels with abnormal, complex shapes which effectively conform to intricate species boundaries.

\subsubsection{Comparison to dense ground-truth masks}
When the DeepLabv3+ architecture is trained on the \emph{dense} ground truth masks provided by the UCSD Mosaics dataset, the performance is 91.11\% for pixel accuracy, 72.22\% for mean pixel accuracy, and 62.86\% for mean intersection over union.  These results can be considered as an upper bound of the achievable performance of this segmentation architecture and training configuration for the UCSD Mosaics dataset.  It is therefore significant to note that our approach, which uses only \emph{300 sparse labels per image} (see Table~\ref{tab:deeplab}) achieves within 2.1\% of the ground truth performance for pixel accuracy, within 4.3\% for mean pixel accuracy and within 4.1\% for mean intersection over union.
\section{Conclusion}
\label{sec:conclusions}
This work has proposed a point label aware method for generating superpixels in underwater imagery.  Unlike prior superpixel approaches, our method leverages the large quantities of sparse expertly-labeled points in generating superpixel segments. Our method improves ground truth augmentation by 3.62\% for pixel accuracy and 8.35\% for mIoU on the UCSD Mosaics dataset.  Our approach also decreases computation time for ground truth augmentation by 76.0\% to give an average of 1.37 seconds per image for our single classifier.  This increase in speed enables faster bootstrapping of models to train on new species whilst in the field.  When training a semantic segmentation model using our augmented ground truth masks, we outperform the state-of-the-art by 2.91\% for pixel accuracy, 8.07\% for mean per class pixel accuracy and 9.65\% for mIoU on the UCSD Mosaics dataset, and by 4.19\% for pixel accuracy, 11.22\% for mean pixel accuracy and 14.32\% for mIoU on the Eilat dataset.

Our point label aware approach to superpixels could be used by ecologists to quickly label new footage collected in the field. A segmentation model can be iteratively improved for the specific geographic location, conditions, species present and based on the interests of the ecologists.  As new species are encountered or as ecological priorities change, the model can learn about these classes without manual curation of a large dataset of densely labeled images.

Our method could be extended by exploiting the spatio-temporal continuity between consecutive frames of a video collected by an AUV or ROV.  Our approach could be initialized with the superpixel decomposition of the previous frame, which could significantly reduce the computation time required to optimize the superpixel centers in the current frame. Propagated labels could be improved using the augmented ground truth from later frames in the sequence, as instances are captured closer to the camera as the vehicle approaches.

Our point label aware approach to superpixels could also be applied to alternate domains such as precision agriculture, specifically for segmentation of weeds for targeted application of herbicides. New weed species could be easily incorporated by providing images accompanied by sparse point labels.

Another avenue for future work is developing a point label aware method for directly training the semantic segmentation network.  This could involve designing a loss function to penalize the model more for mistakes at the location of the point labels and proportionally less the further away the pixel is from the point label, where this distance would be considered both spatially and in the feature space.

\bibliographystyle{IEEEtran} 
\bibliography{IEEEabrv,Bibliography}

\end{document}